\documentclass{article}
\usepackage{log_2025}						

\usepackage{booktabs}						
\usepackage{multirow}						
\usepackage{amsfonts}						
\usepackage{graphicx}						
\usepackage{duckuments}						

\usepackage[numbers,compress,sort]{natbib}	
\usepackage[table]{xcolor}
\usepackage{booktabs}
\usepackage{caption}
\usepackage{makecell} 
\usepackage{algorithm}
\usepackage{algorithmic}
\usepackage{wrapfig}
\usepackage{caption}

\title[Integrating Temporal and Structural Context in Graph Transformers for Relational Deep Learning]{Integrating Temporal and Structural Context in Graph Transformers for Relational Deep Learning}

\author[Lachi et al.]{%
Divyansha Lachi\thanks{Work done during internship at SAP.} \\
University of Pennsylvania \\
Philadelphia, PA, USA \\
\texttt{div11@upenn.edu} \\
\And
Mahmoud Mohammadi \\
SAP \\
Seattle, WA, USA \\
\texttt{mahmoud.mohammadi@sap.com}
\And
Joe Meyer \\
SAP \\
Palo Alto, CA, USA \\
\texttt{joseph.meyer@sap.com}
\And
Vinam Arora \\
University of Pennsylvania \\
Philadelphia, PA, USA \\
\texttt{vinam@upenn.edu} \\
\And
Tom Palczewski \\
SAP \\
Palo Alto, CA, USA \\
\texttt{tom.palczewski@sap.com}
\And
Eva L. Dyer \\
University of Pennsylvania \\
Philadelphia, PA, USA \\
\texttt{eva.dyer@upenn.edu}
}

\usepackage[dvipsnames]{xcolor}      
\usepackage{soul}        
\usepackage{xcolor}      

\newcommand{\hlc}[2][yellow]{ {\sethlcolor{#1} \hl{#2}} }
\iftrue
    \newcommand{\note}[1]{\hlc[Yellow]{\,#1\,}}
    \newcommand{\joe}[1]{\hlc[Apricot]{\textbf{\,Joe:} #1\,}}
    
    \newcommand{\divyansha}[1]{\hlc[GreenYellow]{\textbf{\,Divyansha:} #1\,}}
\else
    \newcommand{\note}[1]{}
    \newcommand{\eva}[1]{}
    \newcommand{\joe}[1]{}
    \newcommand{\divyansha}[1]{}
\fi

\begin{document}
\maketitle
\captionsetup{font=footnotesize}

\begin{abstract}

In domains such as healthcare, finance, and e-commerce, the temporal dynamics of relational data emerge from complex interactions---such as those between patients and providers, or users and products across diverse categories. To be broadly useful, models operating on these data must integrate long-range spatial and temporal dependencies across diverse types of entities, while also supporting multiple predictive tasks. 
However, existing graph models for relational data primarily focus on spatial structure, treating temporal information merely as a filtering constraint to exclude future events rather than a modeling signal, and are typically designed for single-task prediction. To address these gaps, we introduce a temporal subgraph sampler that enhances global context by retrieving nodes beyond the immediate neighborhood to capture temporally relevant relationships. 
In addition, we propose the \textbf{Relational Graph Perceiver (RGP)}, a graph transformer architecture for relational deep learning that leverages a cross-attention-based latent bottleneck to efficiently integrate information from both structural and temporal contexts. This latent bottleneck integrates signals from different node and edge types into a common latent space, enabling the model to build global context across the entire relational system. RGP also incorporates a flexible cross-attention decoder that supports joint learning across tasks with disjoint label spaces within a single model. Experiments on RelBench, SALT, and CTU show that RGP delivers state-of-the-art performance, offering a general and scalable solution for relational deep learning with support for diverse predictive tasks.

\end{abstract}

\section{Introduction}
\label{sec:intro}

Relational data is central to many real-world systems in domains such as healthcare, finance, and e-commerce. These datasets capture interactions between entities such as patients and providers, customers and products, or suppliers and inventory, which unfold over time and span multiple data modalities \cite{codd1970relational}. The data is typically organized in multi-table relational databases and presents a complex modelling challenge, which involves both long-range structural dependencies through entity relationships and temporal dynamics through evolving interactions.

\textbf{Relational Deep Learning (RDL)} provides a principled framework for learning from such data by converting relational databases into heterogeneous temporal graphs \cite{fey2024position}. In this formulation, nodes correspond to entities (e.g., users, items, visits), and edges represent typed relationships (e.g., purchases, interactions, transactions). While traditional RDL models based on Graph Neural Networks (GNNs) have shown success in capturing local structure via message passing, they suffer from several limitations. In particular, GNNs have limited expressiveness \cite{xu2018powerful, jiang2019semi, loukas2019graph} and struggle to capture long-range dependencies due to oversquashing \cite{black2023understanding, qureshi2023limits}.

\textbf{Graph Transformers (GTs)} offer a promising alternative by using attention mechanisms for global aggregation, allowing the model to reason across distant parts of the graph \cite{dwivedi2020generalization, ying2021transformers}. 
RelGT \cite{rgt}, a recent method, adapts GTs to handle the structure and temporal homogeneity found in relational databases.
However, current relational graph models, whether based on GNNs \cite{relbench} or GTs \cite{rgt}, primarily focus on spatial structure and often treat time just as a constraint rather than a modeling signal. In particular, context is sampled around a prediction node where temporal information is typically used to restrict the neighborhood \cite{relbench}, rather than to actively guide the sampling process. 

\noindent\textbf{Our Approach.} 
To address this limitation, we introduce a temporal subgraph sampler that retrieves temporally relevant nodes, allowing the model to reason about nonlocal events that are structurally distant but contextually similar.
To incorporate this rich context, we introduce the \textbf{Relational Graph Perceiver (RGP)}, a graph transformer architecture that efficiently integrates structural and temporal information through a cross–attention–based latent bottleneck. Finally, RGP also supports multi-task learning via a flexible decoder that conditions predictions on task-specific queries and compares them to text-encoded labels using a similarity-based objective.


We evaluate RGP on three diverse benchmarks—\textbf{RelBench}\cite{relbench}, \textbf{CTU}\cite{motl2024ctupraguerelationallearning}, and \textbf{SALT}\cite{klein2025salt}—spanning binary classification, multi-class classification, and ranking-based tasks. RGP consistently achieves strong performance across all settings, while supporting computationally efficient multi-task learning without needing to train separate models or linear layers for each task. These results demonstrate the effectiveness of RGP as a scalable, general-purpose architecture for learning from relational data.

\vspace{0.5em}
\textbf{Our contributions are as follows:}
\begin{itemize}

\item We introduce a novel \textbf{temporal subgraph sampler} that selects nodes based on contextual timestamp proximity, allowing the model to incorporate nonlocal temporal context beyond structural neighborhoods.

\item We present the \textbf{Relational Graph Perceiver (RGP)}, a graph transformer architecture that efficiently integrates structural information and the rich temporal context extracted by our sampler. It uses a cross-attention-based latent bottleneck, enabling scalable global reasoning across heterogeneous relational graphs.

\item We develop a \textbf{flexible multi-task decoder} that enables joint training across diverse tasks with disjoint label spaces. Our decoder uses task-conditioned queries and similarity-based supervision over text-encoded labels, eliminating the need for task-specific output heads.
\end{itemize}


\section{Method}

\begin{figure}[t!]
    \centering
    \includegraphics[width=\linewidth]{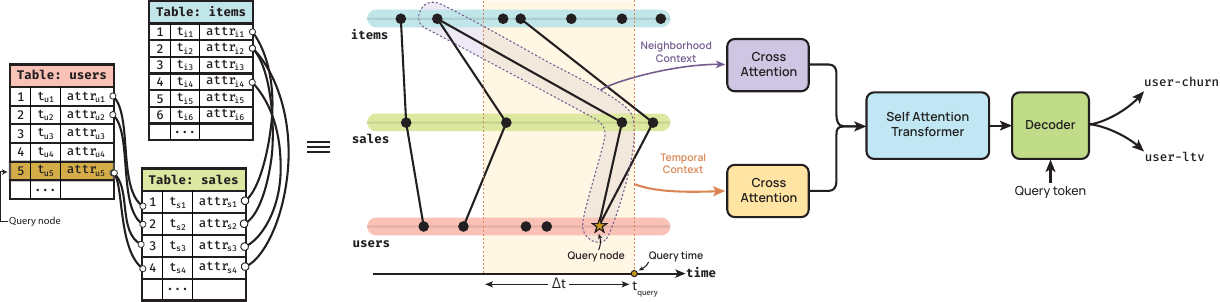}
    \caption{\footnotesize{\textbf{Overview of the RGP architecture.} We convert relational databases into heterogeneous temporal graphs, where nodes represent entities (e.g., users, items, or sales) and edges capture interactions between them. Given a seed or query node (e.g., a user), the model applies two parallel cross-attention modules to encode both structural and temporal context into a set of latent tokens. These latents are then processed by a stack of self-attention transformer blocks to enable long-range reasoning. Finally, a lightweight and flexible decoder maps the latent representation to predictions across multiple tasks, such as user churn or lifetime value (LTV).}\label{fig:overview}} 
    
    
\end{figure}

We now describe the Relational Graph Perceiver (RGP), a general-purpose transformer architecture for learning on heterogeneous temporal relational graphs.
As shown in Figure~\ref{fig:overview}, RGP is built around three key components:
(i) a temporal subgraph sampler that retrieves temporally relevant nodes beyond the immediate neighborhood of the seed or query node (Section~\ref{sec:temporal_sampler});
(ii) a Perceiver-style encoder that uses cross-attention to compress structural and temporal context into a fixed-size latent representation (Section~\ref{sec:encoder}); and
(iii) a lightweight, flexible multi-task decoder that enables training across multiple classification tasks with diverse label spaces, without requiring task-specific linear layers (Section~\ref{sec:decoder}).

Before describing these components in detail, we first explain how relational databases are converted into heterogeneous temporal graphs and tokenized for transformer-based modeling (Section~\ref{sec:tokenization}). The overall pipeline is illustrated in Figure~\ref{fig:overview}, and each component is described in the following sections.


\subsection{Tokenizing Heterogeneous Temporal Graphs}
\label{sec:tokenization}

To process relational data with transformer-based models, we first convert relational databases into graph-structured inputs (Figure~\ref{fig:overview}), enabling end-to-end learning without the need for manual feature engineering. Following prior work in relational deep learning~\cite{rgt, relbench}, we represent relational databases as \textit{relational entity graphs} (REGs),
modeled as heterogeneous temporal graphs.

\paragraph{Relational Graph:}

A relational database can be formally described as a tuple $(\mathcal{T}, \mathcal{R})$, where $\mathcal{T} = {T_1, \dots, T_n}$ is a collection of entity tables, and $\mathcal{R} \subseteq \mathcal{T} \times \mathcal{T}$ is a set of inter-table relationships. Each relation $(T_{\text{fkey}}, T_{\text{pkey}}) \in \mathcal{R}$ denotes a foreign-key reference from one table to the primary key of another. Each table $T_i$ contains a set of entities (rows), where each entity is typically defined by (1) a unique identifier, (2) foreign-key references, (3) entity-specific attributes (e.g., numeric, categorical), and (4) timestamp metadata.


We transform this database into a heterogeneous temporal graph: $\mathcal{G} = (\mathcal{V}, \mathcal{E}, \phi, \psi, \tau)$ where $\mathcal{V}$ is the set of nodes (entities), $\mathcal{E} \subseteq \mathcal{V} \times \mathcal{V}$ is the set of edges (primary-foreign key relationships), $\phi: \mathcal{V} \rightarrow \mathcal{T}_V$ maps each node to its source table (entity type), $\psi: \mathcal{E} \rightarrow \mathcal{T}_E$ assigns relation types to edges, and $\tau: \mathcal{E} \cup \mathcal{V} \rightarrow \mathbb{R}$ associates timestamps with both nodes and edges. This graph representation captures both the schema structure and temporal dynamics of the database.

\paragraph{Token Construction.}
Each node $v_i \in \mathcal{V}$ is mapped to a token embedding $\mathbf{x}_i \in \mathbb{R}^d$ by applying a multi-modal encoder to its raw attributes, followed by the addition of a positional encoding:
\[
\mathbf{x}_i = \texttt{MultiModalEncoder}(\mathbf{u}_i) + \mathrm{PE}(v_i),
\]
where $\mathbf{u}_i$ denotes the raw attributes of the node (e.g., tabular, categorical, or multi-modal features) and \texttt{MultiModalEncoder} is the modality-aware encoder taken from~\cite{hu2024pytorch}. This encoder applies separate encoders for each modality (e.g., numerical, categorical, or text) and aggregates their outputs into a unified embedding using a ResNet (see Appendix~\ref{app:multimodal_encoder} for more details). $\mathrm{PE}(v_i)$ captures structural and temporal context such as node centrality, hop distances, or timestamp embeddings.
Each input graph is mapped to a full input sequence formed by these node-level tokens.

\paragraph{Positional Encodings.}
To represent the position of each node in a heterogeneous and temporal relational graph, we combine multiple structural and time-aware signals into a unified encoding. Specifically, for each node $v_i$, we compute:
\begin{itemize}
    \item \textbf{Node type embedding} $\mathbf{e}_{\text{type}}(v_i)$: a learned embedding based on the node type $\phi(v_i)$.
    \item \textbf{Centrality embedding} $\mathbf{e}_{\text{cent}}(v_i)$: a linear projection of centrality scores (e.g., degree, PageRank).
    \item \textbf{Hop distance embedding} $\mathbf{e}_{\text{hop}}(v_i)$: a learned embedding of the hop distance from a designated entity node (e.g., the seed node or query node in the task).
    \item \textbf{Relative time encoding} $\mathbf{e}_{\text{time}}(v_i)$: a projection of $\tau(v_i) - \tau_{\text{seed}}$ to capture temporal alignment.
\end{itemize}
We concatenate these components and project them into the final positional encoding:
\[
\mathrm{PE}(v_i) = W_{\text{PE}} \cdot \left[ \mathbf{e}_{\text{type}}(v_i) \| \mathbf{e}_{\text{cent}}(v_i) \| \mathbf{e}_{\text{hop}}(v_i) \| \mathbf{e}_{\text{time}}(v_i) \right],
\]
where $W_{\text{PE}} \in \mathbb{R}^{d' \times d}$ is a learned projection matrix and $\|$ denotes concatenation.

\paragraph{Note:}
This multi-element positional encoding allows the model to incorporate fine-grained structural, temporal, and schema-level context across highly diverse relational graphs. While the individual components of this encoding are adapted from established techniques \cite{rgt, ying2021transformers}, their integration provides a unified representation that is suitable for heterogeneous graph modeling.

Once tokenized, we construct an input subgraph around each target entity using both structural sampling \cite{relbench} and temporal context sampling (Section~\ref{sec:temporal_sampler}). When timestamps are missing or partially available, the sampler defaults to standard neighborhood sampling without temporal constraints, allowing the model to operate on both temporal and purely structural relational data. The resulting node sequence is then passed to our Perceiver-based encoder, which compresses it into a fixed-size latent representation via cross-attention (Section~\ref{sec:encoder}).

\subsection{Temporal Sampler}
\label{sec:temporal_sampler}

In many real-world relational systems, temporally correlated events often occur across entities that are not directly connected in the underlying graph.
For instance, a sudden market shift may influence multiple customer accounts that share no explicit transactional links, or a new product release may simultaneously affect several supplier and retail nodes. Capturing information from such temporally proximate but structurally distant entities can provide valuable context for prediction. However, standard neighborhood sampling methods apply time-restricted sampling around a target node to prevent temporal leakage~\citep{rgt, relbench} (Figure~\ref{fig:overview}) by excluding nodes that occur at future timestamp. These methods focus on preserving local graph structure while enforcing temporal constraints, treating time primarily as a boundary condition on graph-based neighborhood sampling.

To complement this, we introduce a second sampling mechanism—the \textit{Time-Context Sampler}, which explicitly leverages temporal proximity as a signal, independent of graph connectivity. This sampler selects edges (and their associated nodes) based on their closeness in time to a reference timestamp, regardless of whether they are direct neighbors of the target node. This enables the model to incorporate temporally co-occurring events that may reflect broader contextual information—such as concurrent user activity, market trends, or environmental conditions—that can be critical for accurate prediction.

Formally, given a graph $G = (V, E)$ with node timestamps $T_v : V \to \mathbb{R}$, we define an edge timestamp $T_e(u,v) = f(T_v(u), T_v(v))$, where $f$ can be the mean or maximum of the endpoint timestamps. Using a seed time $t_{\text{seed}}$, we then extract a temporal subgraph by selecting edges within a fixed time window $\Delta t$ or the $k$ nearest edges in time. For datasets where edge timestamps are already provided, we retain those original values. The full algorithm is provided in Appendix~\ref{app:temporal_sampler_algo}.



\subsection{Compression via Cross Attention}
\label{sec:encoder}

The temporal sampler introduced in Section~\ref{sec:temporal_sampler} provides rich context by retrieving nodes that are temporally relevant but structurally distant. To effectively integrate this expanded context without incurring the quadratic cost of full self-attention, we adopt a Perceiver-inspired encoder~\cite{jaegle2022perceiverio, lachi2024graphfm} that compresses both structural and temporal node embeddings into a fixed set of latent representations through cross-attention.


\vspace{0.3em}
\noindent\textbf{Latent bottleneck.}
Given the tokenized graph structure described in Section~\ref{sec:tokenization} and the subgraph sampled around each target entity (using both structural and temporal samplers), we obtain a sequence of input node embeddings 
$\mathbf{X}_g = [\mathbf{x}_1, \dots, \mathbf{x}_{N_g}]$. 
We maintain a shared set of $K$ \textit{learnable} latent tokens 
$\mathbf{Z}_0 = [\mathbf{z}_{0,1}, \dots, \mathbf{z}_{0,K}]$, 
where $K \!\ll\! N_g$. 
These latent tokens attend to all nodes via a cross-attention mechanism:
\begin{equation}
\label{eq:cross_attn}
\mathbf{Z}_g^{(1)} = 
\mathbf{Z}_0 + 
\mathrm{softmax}\!\left(
    \frac{(\mathbf{W}_q \mathbf{Z}_0)
    (\mathbf{W}_k \mathbf{X}_g)^\top}{\sqrt{d_k}}
\right)
(\mathbf{W}_v \mathbf{X}_g),
\end{equation}
where $\mathbf{W}_q$, $\mathbf{W}_k$, and $\mathbf{W}_v$ are learned projection matrices.
Through this mechanism, the latent tokens integrate information from the entire graph without requiring dense pairwise interactions among all nodes.

\vspace{0.3em}
\noindent\textbf{Fusing temporal and structural context.}
To combine temporal and structural signals, we apply two parallel cross-attention branches—one over temporally sampled nodes and another over structurally sampled nodes.
Each branch maintains its own set of learnable latent tokens and uses an independent cross-attention layer with its own query, key, and value projections. The resulting latent representations from the two branches are summed elementwise to produce a unified latent embedding. This fused latent representation allows the model to integrate information from structurally local and temporally correlated regions of the graph.

Following this compression step, we process the latents with a stack of $L$ self-attention blocks operating purely in the latent space, yielding the final set of compressed latent tokens $\mathbf{Z}_g^{\text{out}}$.  We use standard Transformer blocks with pre-layer normalization and feed-forward layers~\citep{vaswani2017attention}. 
Because self-attention is restricted to $K$ tokens, the computational cost scales as:
\begin{equation}
\mathcal{O}(K N_g + L K^2) \ll \mathcal{O}(N_g^2), \quad K \ll N_g,
\end{equation}
resulting in substantial savings in both compute and memory while still allowing the model to integrate information from all nodes in the input subgraphs.

\subsection{Multi-Task Decoder}
\label{sec:decoder}

\begin{wrapfigure}{r}{0.5\textwidth}
    \centering
   \includegraphics[width=\linewidth]{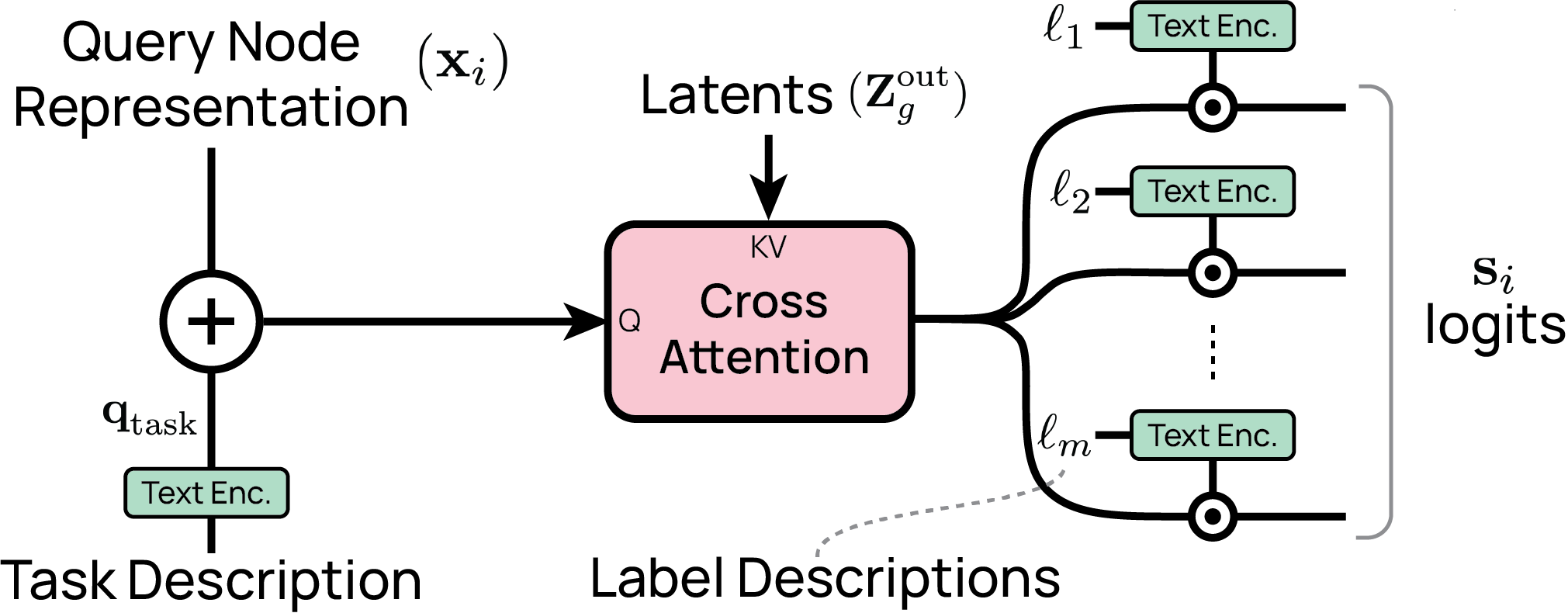}
    \caption{\footnotesize{\textbf{Overview of Flexible multi-task decoder}: The decoder receives a query node representation (combined with a task description embedding) and attends to the latents from the perceiver encoder via cross attention.}}
    \label{fig:multi_task_overview}
    \vspace{-4mm}
\end{wrapfigure}

After the Perceiver encoder compresses the input subgraphs (structural and temporal neighborhood) into a fixed-length latent representation, this representation must be mapped to task-specific predictions. To support diverse prediction objectives across multiple tasks, we adopt a flexible multi-task decoder that combines cross-attention with similarity-based label supervision. Instead of maintaining a separate output head for each task, we use a shared decoding mechanism that conditions predictions on both the task description and the target node representation. This design enables efficient parameter sharing across tasks while allowing task-specific behavior to emerge through the attention mechanism and label embeddings. For each prediction task, we define a task embedding $\mathbf{q}_{\text{task}} \in \mathbb{R}^d$
 that encodes task-specific context (e.g., classification objective or label space). The embedding is obtained by passing the textual task description through a pretrained language model and projecting the resulting representation into the model’s latent space. Given this fixed task embedding and the tokenized representation of the target node $\mathbf{x}_i \in \mathbb{R}^d$ 
(from Section~\ref{sec:tokenization}), we compute a task-aware query via element-wise summation.
\[
\mathbf{q}_i = \mathbf{x}_i + \mathbf{q}_{\text{task}}.
\]

This query is used to attend to the encoder's latent  $\mathbf{Z}^{\text{out}} \in \mathbb{R}^{K \times d}$ using a cross-attention mechanism:

\[
\mathbf{z}_i = \text{CrossAttn}(\mathbf{q}_i, \mathbf{Z}^{\text{out}}_{g}),
\]

where $\mathbf{z}_i \in \mathbb{R}^d$ is the task-conditioned representation of the node.

To enable generalization across tasks without task-specific output layers, we represent each candidate label as a text string and encode it using a frozen or pretrained text encoder. For example, in a user churn prediction task for the \texttt{rel-hm} dataset, a candidate label can be represented as a text string such as ``H\&M customer will stop making transactions in next week," which is then encoded using a pretrained text encoder to obtain the corresponding label embedding.

\[
\mathcal{L} = \{\ell_1, \ell_2, \dots, \ell_m\}, \quad \mathbf{E}_{\text{label}} = \text{TextEncoder}(\mathcal{L}) \in \mathbb{R}^{m \times d},
\]

where $\ell_j$ is the $j$-th label and $m$ is the number of possible labels for the current task.

We compute the logits by taking the dot product between the output node embedding $\mathbf{z}_i$ and each label embedding:

\[
\mathbf{s}_i = \mathbf{E}_{\text{label}} \mathbf{z}_i \in \mathbb{R}^m.
\]

These logits are passed through a softmax function to compute class probabilities, and the loss is computed using standard objectives such as cross-entropy for classification. This unified decoding framework eliminates the need for task-specific classifiers or training separate models for each task, enabling scalable and efficient multi-task learning.

\section{Results}
\label{sec:results}

We evaluate the Relational Graph Perceiver (RGP) on a diverse suite of heterogeneous temporal graph datasets spanning multiple domains and benchmarks. Specifically, we consider three sources: RelBench~\cite{relbench}, CTU~\cite{motl2024ctupraguerelationallearning}, and SALT~\cite{klein2025salt}. Our experiments focus on the node (or entity) classification task, where the goal is to predict categorical attributes associated with entities in relational graphs. As these benchmarks originate from distinct application areas and have only recently been introduced, they differ significantly in terms of available baselines and evaluation metrics. We provide a detailed discussion of the dataset-specific metrics and baseline comparisons in Section~\ref{sec:main_res}.

\subsection{Experimental Setup}

We implement RGP within the RDL pipeline~\cite{relbench} by replacing the original GNN component with our architecture, while preserving the underlying task logic, database loaders, and training infrastructure. RGP is trained using the AdamW optimizer~\cite{loshchilov2017decoupled} with a fixed learning rate of $10^{-3}$. 
Similar to previous work \citep{rgt}, we only tune a few key architectural hyperparameters: total number of layers in the model, $L \in \{2, 4, 6\}$, and the number of latent tokens, $n \in \{8, 16, 32\}$, in the cross-attention block. All other settings, such as batch size and dropout, remain fixed across datasets.  For a complete list of hyperparameters, see Section~\ref{app:hyp} in the appendix. 


\subsection{Results on Benchmarks}
\label{sec:main_res}

We report benchmark results for RGP across three representative datasets—RelBench, CTU, and SALT, each reflecting a different application domain and evaluation metric. Due to the diversity of these benchmarks, we group results by benchmark and compare RGP against the publicly available baselines for each setting. Across all benchmarks, we include comparisons to RDL~\cite{fey2023relational}, a widely adopted pipeline that combines RelGNN~\cite{chen2025relgnn} with GraphSAGE aggregation, serving as a strong reference point for relational deep learning tasks.

\begin{table*}[t]
\centering
\caption{\textbf{Results on Relbench:} We report the Area Under the ROC Curve (AUC) as the evaluation metric. Best values are shown in \textbf{bold}. Relative gains indicate the percentage improvement of RGP over RelGT.}
\label{tab:relbench_auc}
\setlength{\tabcolsep}{10pt}
\renewcommand{\arraystretch}{1.1}
\resizebox{\textwidth}{!}{
\begin{tabular}{l l c c c c c c}
\toprule
Dataset & Task & RDL & HGT & HGT+PE & RelGT & \makecell{\textbf{RGP}\\\textbf{(ours)}} & \makecell{\% Rel Gain \\ vs. RelGT} \\
\midrule
rel-f1     & driver-dnf       & 0.7262 & 0.7142 & 0.7109 & 0.7587 & \textbf{0.7844} & \cellcolor{blue!10}+3.39 \\
           & driver-top3      & 0.7554 & 0.6389 & 0.8340 & 0.8352 & \textbf{0.8789} & \cellcolor{blue!10}+5.22 \\
rel-avito  & user-clicks      & 0.6590 & 0.6584 & 0.6387 & 0.6830 & \textbf{0.6943} & \cellcolor{blue!10}+1.66 \\
           & user-visits      & 0.6620 & 0.6426 & 0.6507 & \textbf{0.6678} & 0.6662 & \cellcolor{red!10}-0.24 \\
rel-event  & user-repeat      & 0.7689 & 0.6717 & 0.6590 & 0.7609 & \textbf{0.7894} & \cellcolor{blue!10}+3.75 \\
           & user-ignore      & 0.8162 & 0.8348 & 0.8161 & 0.8157 & \textbf{0.8439} & \cellcolor{blue!10}+3.46 \\
rel-trial  & study-outcome    & 0.6860 & 0.5679 & 0.5691 & 0.6861 & \textbf{0.7027} & \cellcolor{blue!10}+2.42 \\
rel-amazon & user-churn       & 0.7042 & 0.6608 & 0.6589 & 0.7039 & \textbf{0.7089} & \cellcolor{blue!10}+0.71 \\
           & item-churn       & \textbf{0.8281} & 0.7824 & 0.7840 & 0.8255 & 0.8262 & \cellcolor{blue!10}+0.08 \\
rel-stack  & user-engagement  & 0.9021 & 0.8898 & 0.8818 & \textbf{0.9053} & 0.9045 & \cellcolor{red!10}-0.09 \\
           & user-badge       & \textbf{0.8966} & 0.8652 & 0.8636 & 0.8624 & 0.8868 & \cellcolor{blue!10}+2.83 \\
rel-hm     & user-churn       & 0.6988 & 0.6773 & 0.6491 & 0.6927 & \textbf{0.7025} & \cellcolor{blue!10}+1.41 \\
\midrule
\multicolumn{7}{r}{\textbf{Average Gain vs. RelGT (\%)}} & \cellcolor{blue!10}\textbf{2.20} \\
\bottomrule
\end{tabular}
}
\end{table*}

\vspace{\parskip}
\textbf{Relbench} \cite{relbench} is a recently introduced benchmark for relational deep learning that includes seven datasets derived from structured domains such as e-commerce, social networks, and sports. 
Each dataset has a binary node classification task, where the objective is to predict an entity’s future state or behavior based on its multi-relational and temporal context. For instance, in the \texttt{user-churn} task for \texttt{rel-hm}, the model predicts whether a customer will become inactive (i.e., have no transactions) in the following week. Performance is evaluated using the AUC-ROC metric (refer to Appendix~\ref{app:regression} for results on regression tasks).
For RelBench, we report results from the Relational Graph Transformer (RelGT) \cite{dwivedi2025relational}, the current state-of-the-art method on this benchmark. The paper also compared against HGT \cite{hu2020heterogeneous} and a variation of HGT with Laplacian positional encodings \cite{arora2025exploiting,dwivedi2020generalization}. 
As shown in Table~\ref{tab:relbench_auc}, RGP achieves state-of-the-art performance in RelBench, outperforming RelGT on 10 out of 12 tasks. Notably, RGP yields an average relative improvement of \textbf{2.2\%} over RelGT across all tasks.
Gains are particularly pronounced on smaller datasets, where self-attention-based models like RelGT are more prone to overfitting. For example, on the \texttt{driver-top3} task from the \texttt{rel-f1} dataset RGP outperforms RelGT by \textbf{3.39\%}, and on the \texttt{driver-dnf} task by \textbf{5.22\%}. Similarly, on the \texttt{user-repeat} task from \texttt{rel-event}, we observe a gain of \textbf{3.75\%}.

In terms of computational cost, RGP scales as $O(KN_gd + LK^2d)$ with $K \ll N_g$, achieving near-linear dependence on the number of input nodes $N_g$. In contrast, HGT and RelGT require full self-attention with $O(N_g^2d)$ complexity, quadratic in input size. For comparison, message-passing GNNs operate in $O(M_gd)$ time, where $M_g$ denotes the number of edges. RGP therefore offers a favorable trade-off between efficiency and expressivity, maintaining transformer-level performance while significantly reducing computational overhead.





\begin{table*}[t]
\centering
\scriptsize 
\setlength{\tabcolsep}{3.8pt} 
\renewcommand{\arraystretch}{1.1}
\caption{\textbf{Results on CTU benchmarks:} We report F1 score as the evaluation metric. Best values are in \textbf{bold}. Relative gains denote percentage improvement of RGP over DBFormer and LightGBM.}
\label{tab:ctu}
\resizebox{\textwidth}{!}{
\begin{tabular}{l l c c c c c c c c c}
\toprule
Dataset & Task & LightGBM & XGBoost & TabResNet & Linear & \makecell{SAGE\\(RDL)} & DBFormer & \makecell{\textbf{RGP}\\\textbf{(ours)}} & \makecell{Rel.\\(\%) Gain\\vs. DBF} & \makecell{Rel.\\(\%) Gain\\vs. LGBM} \\
\midrule
accidents & temp. &
0.170 & 0.336 & 0.187 & 0.583 & 0.566 & 0.727 & \textbf{0.743} &
\cellcolor{blue!10}+2.20 & \cellcolor{blue!10}+337.06 \\

dallas & temp. &
\textbf{0.584} & 0.512 & 0.247 & 0.393 & 0.424 & 0.513 & 0.555 &
\cellcolor{blue!10}+8.19 & \cellcolor{red!10}-4.96 \\

legalacts & temp. &
\textbf{0.851} & 0.220 & 0.220 & 0.721 & 0.698 & 0.703 & 0.736 &
\cellcolor{blue!10}+4.69 & \cellcolor{red!10}-13.52 \\
\bottomrule
\end{tabular}
}
\vspace{-5mm}
\end{table*}

\vspace{\parskip}
\textbf{CTU} \cite{motl2024ctupraguerelationallearning} is a curated repository of heterogeneous graph datasets from domains such as insurance, law and retail. We evaluate on CTU because it includes multi-class classification tasks, enabling us to test RGP beyond binary settings.
For example, in the \texttt{LegalActs} dataset, the model predicts the \texttt{ActKind} (type of court decision) based on case metadata, associated judges, and related legal documents.
We compare against baselines reported in the ReDeLEx benchmark~\cite{pelevska2025redelex}, which include traditional tabular models (e.g., LightGBM \cite{ke2017lightgbm}, XGBoost \cite{chen2016xgboost}), temporal MLPs (e.g., TabResNet \cite{gorishniy2021revisiting}), and the GNN-based RDL \cite{relbench} framework. Following standard practice from ReDeLEx \citep{pelevska2025redelex}, we report F1 scores for all tasks.
As shown in Table~\ref{tab:ctu}, RGP consistently outperforms the DBFormer baseline across all evaluated CTU datasets. Notably, on the \texttt{dallas} dataset, RGP achieves a relative gain of \textbf{8.19\%}. However, we note that tree-based methods such as LightGBM outperform RDL-based approaches on two out of three CTU tasks. This observation aligns with prior work \citep{grinsztajn2022tree,pelevska2025redelex}, which shows that boosting-based models tend to perform well on smaller or flatter relational databases with few one-to-many links but many factual (non-key) columns. LightGBM performs strongly on \textit{Dallas} and \textit{LegalActs} because most predictive information in these datasets is contained within a single table composed primarily of categorical and factual metadata. In contrast, \textit{Accidents} links each event to multiple participants with distinct roles and attributes, where outcomes depend on their joint interactions and contextual factors such as location and time. Flattening these relationships into a single table discards this relational structure, limiting the ability of tree-based methods to model inter-entity dependencies. Models like RGP, which explicitly capture cross-entity and temporal relationships, are therefore better suited for such settings.


\vspace{\parskip}
\textbf{SALT} 
\cite{motl2024ctupraguerelationallearning} is a 
real-world dataset derived\hfill\\[-4mm]
\begin{wraptable}{r}{0.55\textwidth}
\centering
\scriptsize
\setlength{\tabcolsep}{5pt}
\renewcommand{\arraystretch}{1.1}
\vspace{-8mm}
\caption{
\textbf{Results on SALT:} We report MRR score as the evaluation metric. Best values are in \textbf{bold}. Relative gains are percentage improvement over HGT.}
\label{tab:salt}
\begin{tabular}{l c c c c}
\toprule
Task & RDL & HGT & \makecell{\textbf{RGP}\\\textbf{(ours)}} & \% Rel Gain v.s HGT \\
\midrule
item-incoterms  & 0.64 & 0.75 & \textbf{0.81} & \cellcolor{blue!10}+8.00 \\
sales-group     & 0.20 & 0.31 & \textbf{0.34} & \cellcolor{blue!10}+9.68 \\
sales-payterms  & 0.39 & \textbf{0.60} & 0.58 & \cellcolor{red!10}-3.33 \\
sales-shipcond  & 0.59 & 0.76 & \textbf{0.81} & \cellcolor{blue!10}+6.58 \\
\bottomrule
\end{tabular}
\end{wraptable} %
from an Enterprise Resource Planning (ERP) system.
It consists of ranking-based entity classification tasks that reflect practical industrial decision-making scenarios. Since these tasks involve ranking objectives, we report Mean Reciprocal Rank (MRR) as the evaluation metric. For example, in the \texttt{sales-shipcond} task, the model predicts the \texttt{ShippingCondition} of an order, such as the applicable delivery terms, given information about the customer, product, and sales document context. As graph-based models have not been previously applied to SALT, we compare RGP against standard baselines, including RDL and the Heterogeneous Graph Transformer (HGT)~\cite{hu2020heterogeneous}. 
As shown in Table~\ref{tab:salt}, RGP frequently outperforms both baselines for three out of four tasks. For the \texttt{sales-payterms} task, our model is slightly below HGT (0.58 vs. 0.60 MRR) while still outperforming RDL.

\vspace{-1mm}
\paragraph{Summary:}Across all three benchmarks, RGP demonstrates robust performance gains over strong baselines in diverse settings—binary classification (RelBench), multi-class classification (CTU), and ranking-based tasks (SALT). These results collectively highlight RGP’s versatility and effectiveness as a general-purpose relational graph model.


\subsection{Ablation Studies}

\begin{figure}[t!]
    \centering
    \includegraphics[width=\linewidth]{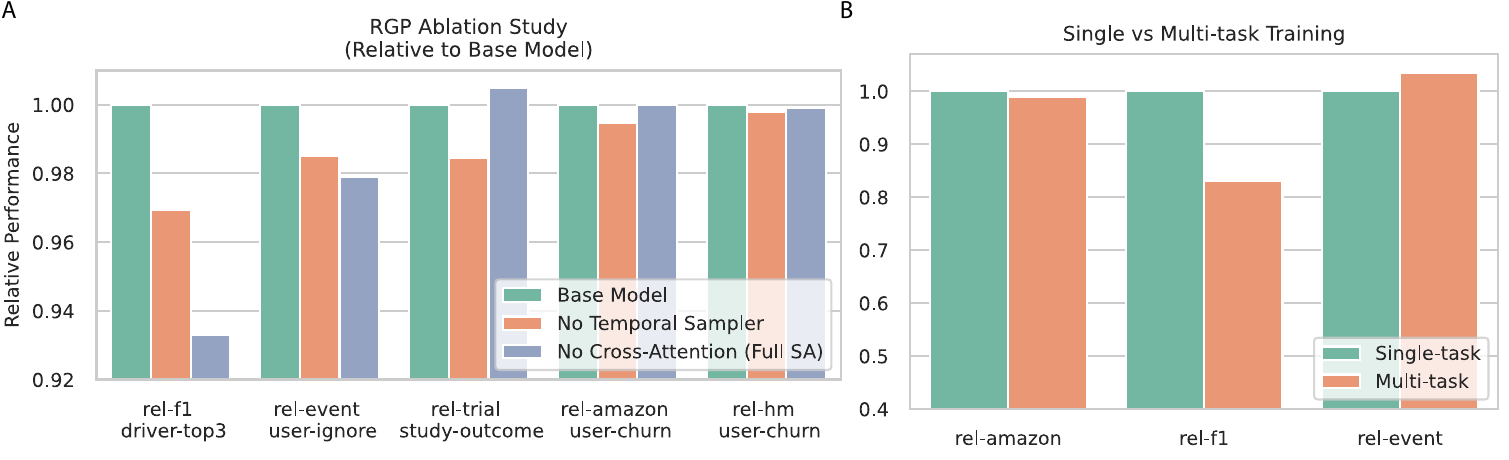}
    \caption{\footnotesize{\textbf{(A)} Results from ablation study of RGP. We evaluate the impact of removing key components from the full RGP model. Performance is reported relative to the base model. A decrease in performance indicates that the removed component is important to overall model effectiveness.
    \textbf{(B)} Relative performance of multi-task vs. single-task training across datasets: the Y-axis shows the average multi-task performance normalized with respect to the average single-task performance across all tasks within each dataset. }\label{fig:ablation}}

\end{figure}

To better understand the contributions of key architectural components in RGP, we perform a series of ablation studies. Specifically, we analyze the impact of (1) removing the temporal context sampler and (2) replacing the Perceiver-style cross-attention bottleneck with full self-attention. For both settings, we report the relative performance compared to the best full model in Figure~\ref{fig:ablation}A.

\paragraph{Temporal sampler:} The temporal sampler is designed to retrieve nodes based on temporal proximity, complementing the structural neighborhood sampler. As shown in Figure~\ref{fig:ablation}A, removing the temporal sampler leads to consistent performance drops, most notably on the \texttt{rel-f1} dataset, where we observe a $\sim$3\% decline. This dataset contains many cold-start cases, such as newly introduced driver nodes with limited or no structural history. In such cases, structural context alone is insufficient for reliable predictions.

\paragraph{Perceiver encoder:}We next evaluate the role of the latent Perceiver encoder by replacing it with a full self-attention (SA) mechanism applied over the input tokens. As shown in Figure~\ref{fig:ablation}A, this substitution yields marginal improvements on a few tasks but at the cost of significantly more compute requirements. Additionally, for smaller datasets such as \texttt{rel-f1}, full self-attention leads to overfitting and degraded performance, with a $\sim$6.6\% drop on the \texttt{driver-top3} task. In contrast, the cross-attention bottleneck in RGP offers a more compute-efficient and regularized alternative, preserving competitive performance while maintaining scalability across tasks. On average, cross-attention is 2–6x more compute-efficient in comparison to self-attention (see Appendix~\ref{app:compute_time} for run times).



\subsection{Multi-Task Results}


A key motivation behind the flexible decoder introduced in Section~\ref{sec:decoder} is to enable scalable multi-task learning across diverse label spaces. Most existing approaches for relational graph learning adopt a task-specific training paradigm, where separate models are trained for each task, even when these tasks share the same underlying graph structure. 

In contrast, the decoder in RGP supports task-conditioned decoding by combining cross-attention with a similarity-based objective over text-encoded label embeddings. This unified framework allows a single model to learn multiple tasks simultaneously.
%
To evaluate the effectiveness of this design, we compare single-task training (one model per task) with multi-task training (a single model trained jointly on all tasks from the same dataset). In our multi-task setting, task supervision is provided via text-based task queries and label embeddings, as described in Section~\ref{sec:decoder}. 

As shown in Figure~\ref{fig:ablation}B, RGP achieves comparable or improved results under multi-task training. For the event dataset, multi-task training even led to slight improvements over single-task models. However, we observed a notable drop in performance on the \texttt{rel-f1} dataset under the multi-task setting. This can be attributed to the severe imbalance in the number of training samples across tasks: the \texttt{driver-top3} task has only 
$\sim$1.5k samples, whereas \texttt{driver-dnf} has over 10k. Such imbalance likely causes the shared model to underfit the smaller task, leading to degraded performance on \texttt{driver-top3}.

Overall, these results highlight the generalization ability of our decoder and its parameter-efficient multi-task learning without any retraining or architectural modification, making it a strong candidate for large-scale foundation model training across relational datasets.

\section{Related Work}

Representing relational datasets as heterogeneous temporal graphs has enabled the use of graph learning methods for tasks like node classification and link prediction. The RelBench benchmark~\cite{relbench} formalizes this paradigm and provides a strong baseline using Heterogeneous GraphSAGE \cite{Hamilton:2017tp} with temporal neighbor sampling, surpassing classical tabular methods like LightGBM \cite{ke2017lightgbm}. Several architectures have been proposed to better exploit relational structure: RelGNN \cite{chen2025relgnn} introduces composite message passing to preserve information across bridge and hub nodes, while ContextGNN \cite{yuan2024contextgnn} combines pair-wise and two-tower encoders for recommendation scenarios.

Graph Transformers (GTs) adapt the self-attention paradigm \cite{vaswani2017attention} to graph-structured data, capturing long-range dependencies without iterative neighborhood aggregation \cite{dwivedi2021generalization}. Early GT variants focused on local attention with positional encodings like Laplacian eigenvectors \cite{dwivedi2022lspe}, while later designs incorporated global attention mechanisms \cite{ying2021transformers, mialon2021graphit,kreuzer2021rethinking} and scaling strategies such as hierarchical pooling or sparse attention. Heterogeneous GTs such as HGT \cite{hu2020heterogeneous} and Hinormer \cite{mao2023hinormer} model multi-type graphs, but face quadratic cost and per-task training inefficiency. Most relevant to our work is the Relational Graph Transformer (RelGT) \cite{rgt}, which introduced a hybrid local/global attention, establishing strong performance on RelBench. Our approach differs by adopting a Perceiver-style latent bottleneck and temporal sampling, enabling greater scalability and multi-task capability.




While GTs highlight the importance of global context, their computational footprint motivates exploring efficiency-focused alternatives. The Perceiver architecture \cite{jaegle2021perceiver} introduces a fixed-size latent array that attends to high-dimensional inputs via cross-attention, followed by latent self-attention, decoupling input size from computational complexity. Perceiver IO \cite{jaegle2022perceiverio} extends this framework to handle diverse output domains from a shared latent representation. In graph learning, early Perceiver-inspired adaptations have been proposed for homogeneous graphs or static settings \cite{jiang2015defining, lachi2024graphfm}, but these typically omit temporal sampling and are not optimized for multi-task inference. Our encoder adapts this latent bottleneck principle specifically to heterogeneous temporal graphs.


Existing RDL methods excel at modeling relational structure but face oversquashing and limited temporal reach. Graph Transformers capture long-range context but remain computationally heavy and often schema-restrictive. Perceiver-based models address scalability but have not been tailored to heterogeneous temporal graphs or multi-task learning in relational settings. This motivates architectures like ours that combine the latent efficiency of Perceivers with relational and temporal inductive biases, while enabling shared encoders across multiple tasks.

\section{Conclusion}

We introduced a temporal subgraph sampler that expands the relevant input context by retrieving nodes that are temporally proximate yet structurally distant, enabling the incorporation of context that traditional neighborhood sampling fails to capture. 
Building on this, we proposed the Relational Graph Perceiver (RGP)—a transformer-based architecture that efficiently integrates temporal and structural information via a Perceiver-style cross-attention bottleneck, and supports multi-task prediction across diverse label spaces within a single model. Across multiple benchmarks, RGP consistently outperforms strong baselines, achieving state-of-the-art results on both binary and multi-class tasks while reducing computational overhead. Our ablation studies further validate the importance of each component, particularly the importance of temporally aligned but structurally distant context in improving predictive performance. Beyond accuracy gains, RGP supports flexible multi-task learning without requiring task-specific output heads, positioning it as a strong candidate for large-scale foundation models in relational domains.

\section*{Acknowledgements}
Thanks to Shivashriganesh P. Mahato and Keertika Saroj for insightful discussions and feedback on this manuscript. This project was also supported by NSF CAREER Award RI:2146072, NSF
award CIF:RI:2212182 as well as generous gifts from the CIFAR Azrieli Global Scholars Program (D.L, V.A and E.L.D).


\bibliographystyle{unsrtnat}
\bibliography{reference}

\newpage
\appendix
\section{Appendix}

\subsection{Model Details}
\subsubsection{Multi-modal Encoder}
\label{app:multimodal_encoder}

In heterogeneous temporal relational graphs derived from relational databases, each node may contain a rich set of multi-modal attributes, such as numerical values, categorical fields, text descriptions, timestamps, and image features. To obtain expressive node-level representations suitable for downstream tasks, we use a modality-aware feature encoder taken from prior work in relational deep learning~\cite{relbench,hu2024pytorch}.

Given a node $v \in \mathcal{V}$, we denote its raw attributes as $\mathbf{x}_v = \{\mathbf{x}_v^{(m)}\}_{m \in \mathcal{M}_v}$, where $\mathcal{M}_v \subseteq \mathcal{M}$ is the subset of modalities present for node $v$. Each feature within each modality is independently encoded using a modality-specific function $\phi_m$, yielding a set of intermediate embeddings:
\begin{align}
\mathbf{h}_v^{(m)} &= \phi_m\left( \mathbf{x}_v^{(m)} \right), \quad \phi_m: \mathcal{X}^{(m)} \to \mathbb{R}^{d_m},
\end{align}
where $d_m$ is the dimensionality assigned to modality $m$. Supported modalities include:
\begin{itemize}
    \item \textbf{Numerical features:} encoded via linear layers or small MLPs.
    \item \textbf{Categorical features:} embedded using learned lookup tables.
    \item \textbf{Text features:} embedded via pretrained or fine-tuned language models (e.g., BERT).
    \item \textbf{Timestamps:} embedded as scalar values or periodic functions (e.g., sinusoidal encodings).
\end{itemize}

The resulting embeddings are concatenated:
\begin{align}
\mathbf{h}_v^{\text{concat}} &= \bigoplus_{m \in \mathcal{M}_v} \mathbf{h}_v^{(m)} \in \mathbb{R}^{\sum_{m \in \mathcal{M}_v} d_m},
\end{align}
and passed through a table-specific (or node-type-specific) projection function—a ResNet in our case—to obtain the final node representation:
\begin{align}
\mathbf{h}_v^{(0)} &= f_{\tau(v)}\left( \mathbf{h}_v^{\text{concat}} \right), \quad f_{\tau(v)}: \mathbb{R}^{\sum d_m} \to \mathbb{R}^{d},
\end{align}
where $\tau(v)$ denotes the node type (i.e., source table) and $d$ is the unified hidden dimension used across the model.

\subsubsection{Time-Context Sampling Algorithm}
\label{app:temporal_sampler_algo}

\paragraph{Edge Timestamp Assignment.}
Each edge is assigned a timestamp equal to the maximum timestamp of its two endpoint nodes:
\[
T_e(u, v) = \max\left(T_v(u), T_v(v)\right).
\]

\paragraph{Time-Context-Aware Sampling.}
Given a reference timestamp $t_{\text{seed}}$, we sample a subgraph by selecting edges based on one of two strategies:
\begin{itemize}
    \item Time window: include all edges where $|T_e(u,v) - t_{\text{seed}}| \leq \Delta t$.
    \item Top-$k$: select the $k$ edges closest in time to $t_{\text{seed}}$.
\end{itemize}
The resulting subgraph contains all nodes incident to the selected edges.

In practice, the edge-sorting step required for sampling is performed once during preprocessing. During training and inference, sampling reduces to simple lookups within these pre-sorted adjacency lists, making its computational cost negligible compared to attention operations.

\begin{algorithm}[H]
\caption{Time-Context-Aware Edge Sampling}
\label{alg:time_context}
\begin{algorithmic}[1]
\REQUIRE Graph $G = (V, E)$, timestamp function $T : V \rightarrow \mathbb{R}$, reference time $t_{\text{seed}}$, sampling rule: time window $\Delta t$ or edge count $k$
\ENSURE Subgraph $G_{\text{sub}} = (V_{\text{sub}}, E_{\text{sub}})$

\STATE \textbf{Initialize} $E_{\text{scored}} \gets \emptyset$
\FOR{each edge $(u, v) \in E$}
    \STATE $t_e \gets \max(T(u), T(v))$
    \STATE $\text{score} \gets |t_e - t_{\text{seed}}|$
    \STATE Add $(u, v, t_e, \text{score})$ to $E_{\text{scored}}$
\ENDFOR

\IF{time window $\Delta t$ is specified}
    \STATE $E_{\text{selected}} \gets \{(u, v) ~|~ (u, v, t_e, s) \in E_{\text{scored}},~ s \leq \Delta t\}$
\ELSIF{edge count $k$ is specified}
    \STATE Sort $E_{\text{scored}}$ in ascending order by score
    \STATE $E_{\text{selected}} \gets$ first $k$ edges from sorted list
\ELSE
    \STATE $E_{\text{selected}} \gets E$
\ENDIF

\STATE $V_{\text{sub}} \gets$ nodes appearing in $E_{\text{selected}}$
\RETURN $G_{\text{sub}} = (V_{\text{sub}}, E_{\text{selected}})$
\end{algorithmic}
\end{algorithm}


\subsection{Experiment details}
\subsubsection{Hyperparameter}
\label{app:hyp}

We use a controlled hyperparameter setup to ensure consistent evaluation across diverse datasets without exhaustive tuning. Following prior work, we fix most settings and tune only two architectural hyperparameters: the number of Perceiver layers $L \in \{2, 4, 6\}$ and the number of latent tokens $n \in \{8, 16, 32\}$ in the cross-attention bottleneck. All other hyperparameters are kept fixed across datasets to reflect realistic training scenarios where compute budgets limit per-task tuning.

Table~\ref{tab:hyp} summarizes the fixed hyperparameter settings used in our experiments. All models are trained using the AdamW optimizer with a learning rate of $10^{-3}$ and a weight decay of $10^{-5}$. We use a cosine learning rate scheduler with 10 warmup steps, and all experiments are run for 200 epochs. All experiments were conducted on a single NVIDIA B200 GPU.

\begin{table}[h]
\centering
\caption{Summary of hyperparameter settings used in RGP experiments.}
\label{tab:hyp}
\begin{tabular}{ll}
\toprule
\textbf{Component} & \textbf{Value / Setting} \\
\midrule
\multicolumn{2}{l}{\textbf{Optimizer and Training Setup}} \\
\midrule
Optimizer & AdamW \\
Learning rate & $10^{-3}$ \\
Weight decay & $10^{-5}$ \\
Scheduler & Cosine with 10 warmup steps \\
Epochs & 200 \\
Batch size & 512 \\
\midrule
\multicolumn{2}{l}{\textbf{Model Architecture}} \\
\midrule
Latent token count $n$ & \{8, 16, 32\} (tuned) \\
Transformer layers $L$ & \{2, 4, 6\} (tuned) \\
Dropout & 0.2 \\
hidden dim & 128 \\
\midrule
\multicolumn{2}{l}{\textbf{Temporal Sampler}} \\
\midrule
Edges per type & 10 \\
Temporal decay & 0.1 \\
\bottomrule
\end{tabular}
\end{table}

\subsubsection{Runtime Comparison: Cross-Attention vs Self-Attention}
\label{app:compute_time}

To quantify the efficiency benefits of using a cross-attention (CA) bottleneck over full self-attention (SA), we measure the total training time across three representative tasks. As shown in Table~\ref{tab:runtime}, the Perceiver-based encoder with CA consistently achieves lower runtime compared to its SA counterpart. While SA sometimes offers marginal accuracy improvements on larger datasets, the increase in computational cost is substantial.

\begin{table}[h]
\centering
\caption{Average training time for Cross-Attention (CA) vs. Self-Attention (SA) models. Experiments were run on a single B200 GPU.}
\label{tab:runtime}
\begin{tabular}{lcc}
\toprule
\textbf{Dataset} & \textbf{Cross-Attention (CA)} & \textbf{Self-Attention (SA)} \\
\midrule
\textit{rel-f1}        & 2.7 min   & 7.5 min \\
\textit{rel-amazon}    & 3.5 hrs   & 18.5 hrs \\
\textit{rel-event}     & 4.5 min   & 22 min \\
\bottomrule
\end{tabular}
\end{table}

\subsection{Additional Results}

\subsubsection{Performance on Relbench regression tasks}
\label{app:regression}

As shown in Table~\ref{tab:relbench_regression}, RGP consistently outperforms baseline methods across all evaluated datasets, with particularly strong performance on the Avito dataset. 

\begin{table*}[h]
\centering
\caption{\textbf{Results on regression tasks for Relbench:} We report Mean Absolute Error (MAE, lower is better). Best values are shown in \textbf{bold}.}
\label{tab:relbench_regression}
\setlength{\tabcolsep}{6pt}
\renewcommand{\arraystretch}{1.1}
\begin{tabular}{l l c c c c c}
\toprule
Dataset & Task & RDL & HGT & HGT+PE & RelGT & \makecell{\textbf{RGP}\\\textbf{(ours)}} \\
\midrule
F1      & driver-position & 4.022  & 4.1598 & 4.2358 & 3.9170 & \textbf{3.8702} \\
avito   & ad-ctr          & 0.0410 & 0.0441 & 0.0494 & 0.0345 & \textbf{0.01166} \\
hm      & item-sales      & 0.0560 & 0.0655 & 0.0641 & 0.0536 & \textbf{0.05269} \\
\bottomrule
\end{tabular}
\end{table*}

\subsubsection{Effect of Latent Token Count}

\begin{wrapfigure}{r}{0.55\textwidth}
    \centering
   \includegraphics[width=\linewidth]
   {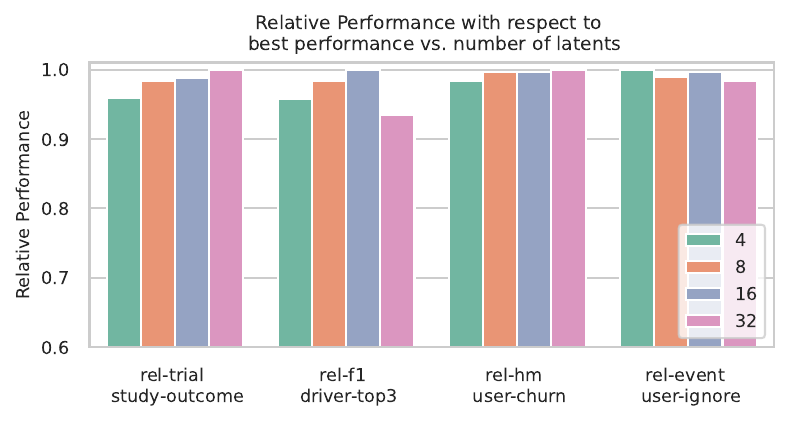}
    \caption{\footnotesize{Effect of number of latent tokens on model performance across four representative tasks. For each task, we normalize results with respect to the best-performing configuration to compute relative performance.}}
    \label{fig:ablation_latent_num}
    \vspace{-3mm}
\end{wrapfigure}

We examine how the number of latent tokens affects performance. We sweep across $n \in \{4, 8, 16, 32\}$ latent tokens on four representative tasks: \textit{study-outcome}, \textit{driver-top3}, \textit{user-churn}, and \textit{user-ignore}. 
For each task, we normalize results with respect to the best-performing configuration to compute relative performance. As shown in Figure~\ref{fig:ablation_latent_num}, RGP achieves strong performance even with a relatively small number of latent tokens. However, the optimal number of latents varies across tasks and datasets, and increasing the number does not always lead to better results. This is why we chose to keep it as a tunable hyperparameter.
For example, the performance of \textit{driver-top3} peaks at 16 latents and declines at 32, suggesting potential overfitting.

\newpage

\subsubsection{Multi-Task Results}
\label{app:multi_task}

\begin{wrapfigure}{r}{0.5\textwidth}
    \centering
    \vspace{-8mm}
   \includegraphics[width=\linewidth]{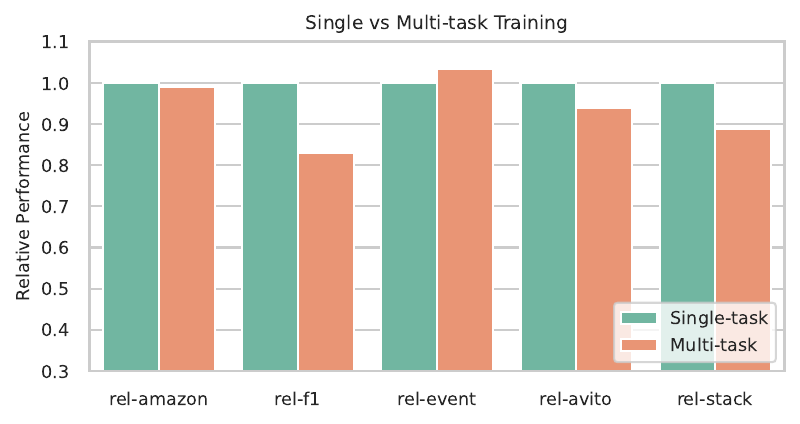}
    \caption{\footnotesize{Relative performance of multi-task vs. single-task training across datasets: The Y-axis shows the average multi-task performance normalized with respect to the average single-task performance across all tasks within each dataset.}}
    \label{fig:multi_task_all}
    \vspace{-6mm}
\end{wrapfigure}

We compare RGP’s performance under single-task and multi-task training across all datasets to assess the scalability of the proposed decoder. In the multi-task setup, a single model is trained jointly on all tasks within each dataset using task-conditioned supervision via text-based task queries and label embeddings.

As shown in Figure~\ref{fig:multi_task_all}, RGP maintains comparable or improved performance under multi-task training in most datasets. While the \emph{event} dataset benefits from joint training, the \emph{f1} dataset shows a slight drop due to task imbalance. Overall, these results demonstrate that RGP’s decoder supports efficient and generalizable multi-task learning across diverse relational tasks.

\end{document}